\documentclass[10pt]{article}
\usepackage[preprint]{tmlr}
\usepackage{graphicx}
\usepackage{amsmath,amssymb,booktabs}
\usepackage{hyperref}
\usepackage{url}
\title{Efficient Constrained Graph Search for Post-hoc Error Correction in Binary Classifiers}
\author{\name Qinwu Xu \email qinwu.xu@utexas.edu\\\addr Leidos Holdings Inc.; The University of Texas at Austin}
\def\month{MM}\def\year{YYYY}\def\openreview{\url{https://openreview.net/forum?id=XXXX}}
\begin{document}
\maketitle
\begin{abstract}
We introduce a model-agnostic framework for constrained post-hoc error correction in binary classifiers. Given a frozen base classifier, the method discovers interpretable conjunctions of feature--threshold conditions that reduce residual false-positive or false-negative errors while limiting adverse changes to previously correct predictions. The proposed approach combines multi-path graph search with depth-dependent constraint tightening, allowing broad candidate correction regions to be progressively refined toward feasible solutions. A reduced-histogram procedure further accelerates threshold evaluation. Unlike classifier retraining or conventional rule learning, the framework optimizes a single correction path over the residual errors of an existing classifier without modifying its parameters. Experiments on a large binary-classification problem demonstrate substantial error reduction, including a configuration that removes 90\% of false positives while sacrificing 5\% of true positives. Additional evaluation across eight OpenML classification tasks demonstrates improvements in the targeted error type on seven of eight tasks in each correction mode, while also revealing the importance of held-out constraint validation.
\end{abstract}

\section{Introduction}

Modern classification systems---including support vector machines (SVMs),
tree ensembles, boosting methods, and neural networks---can achieve strong
aggregate predictive performance while retaining systematic residual errors
in operationally important regions of the input space. In many applications,
the costs of different errors are asymmetric: for example, reducing false
positives (FPs) may be desirable only if the resulting loss of true positives
(TPs) remains below an acceptable tolerance. Cost-sensitive learning has long
studied classification under unequal misclassification costs
\citep{elkan2001}, while selective classification studies the related problem
of trading prediction coverage for reduced predictive risk
\citep{elyaniv2010,geifman2017,geifman2019}. These approaches establish the
importance of explicitly controlling prediction trade-offs rather than
optimizing aggregate accuracy alone. More broadly, optimization methods range from Newton-type numerical
optimization for continuous physical inverse problems
\citep{xu2014,xu2016} to discrete search over structured decision spaces,
with the present work addressing the latter setting.

A complementary problem arises when an already-trained classifier performs
well overall but exhibits identifiable residual errors. Retraining or replacing
the underlying model may be undesirable because of computational cost,
deployment constraints, certification requirements, or limited access to model
parameters. Classifier patching provides one related approach: rather than
replacing a strong but fixed black-box classifier, an auxiliary model can
identify regions in which the original classifier is error-prone and modify
its behavior in those regions \citep{kauschke2018}. Recent work on post-hoc
selective classification likewise studies improvements that can be applied
to pretrained classifiers without modifying their underlying parameters
\citep{cattelan2024}. These developments motivate a broader question: given a
frozen binary classifier, can a compact and interpretable post-hoc rule remove
a targeted subset of residual errors while explicitly limiting adverse changes
to predictions that were already correct?

We formulate this problem as \emph{constrained post-hoc error correction}.
Let the frozen base classifier be
\begin{equation}
    \hat{y}=C_{\theta}(x),
\end{equation}
where $\theta$ is fixed. We learn a correction operator
\begin{equation}
    \hat{y}'=R_{\phi}(x,\hat{y}),
\end{equation}
where $\phi$ specifies a conjunction of feature identities, thresholds,
inequality directions, and their top-down rule path. The base-model parameters
$\theta$ are never modified; optimization is restricted to the post-hoc rule
parameters $\phi$.

Our formulation is also related to interpretable tree and rule learning.
Decision trees recursively partition the feature space through simple
feature-threshold decisions \citep{quinlan1987}. Rule-based approaches instead
represent predictions using compact logical conditions or decision lists.
Examples include predictive rule ensembles \citep{friedman2008}, Bayesian rule
lists \citep{letham2015}, and certifiably optimal rule lists
\citep{angelino2018}. More recent work has addressed limitations of greedy
tree induction through explicit optimization or alternative interpretable
tree structures, including optimal sparse decision trees
\citep{hu2019} and Fast Interpretable Greedy-Tree Sums (FIGS)
\citep{tan2025}. These methods demonstrate both the usefulness of compact
rule-based models and the limitations of committing prematurely to locally
greedy splits.

The objective considered here is different from learning a complete
interpretable classifier from scratch. We assume that $C_{\theta}$ already
exists and restrict the search to a post-hoc correction of its predictions.
Rather than selecting one locally optimal split and discarding alternatives,
the proposed method retains multiple high-ranked candidate
feature--threshold rules and organizes them into a graph. Candidate paths are
explored using depth-first search (DFS) with top-$K$ pruning. A depth-dependent
dynamic optimization schedule permits broader correction at upper levels while
progressively enforcing the final FP or TP constraint. A reduced-histogram
procedure further reduces the cost of searching candidate thresholds.

\paragraph{Contributions.}
Our main contributions are:
(1) a model-agnostic formulation of constrained post-hoc error correction for
frozen binary classifiers, explicitly separating the base-model parameters
$\theta$ from the learned correction-rule parameters $\phi$;
(2) a graph-search procedure that retains and explores multiple candidate rule
paths rather than committing to a single greedy split;
(3) a depth-dependent dynamic optimization strategy that progressively enforces
the desired FP--TP trade-off as search depth increases; and
(4) a reduced-histogram threshold-search procedure designed to reduce
computational cost while retaining greater resolution around informative
thresholds.

\section{Related Work and Motivation}

\subsection{Decision Trees and Interpretable Rule Learning}

Decision trees are among the most established interpretable classification
models. Starting from a root node, they recursively partition the input space
using feature-dependent decisions until a terminal prediction is reached
\citep{quinlan1987}. Their hierarchical structure provides a direct
representation of the conditions leading to a prediction, but standard tree
induction is generally greedy: a locally selected split can restrict the
quality of subsequent paths.

A substantial literature has therefore explored alternative forms of
interpretable rule learning. RuleFit constructs predictive models from linear
combinations of simple rules, where individual rules are conjunctions of a
small number of feature conditions \citep{friedman2008}. Bayesian Rule Lists
learn sparse ordered decision lists intended to remain understandable to human
users \citep{letham2015}. CORELS uses discrete optimization and algorithmic
bounds to construct sparse rule lists with certificates of optimality
\citep{angelino2018}. Similarly, Optimal Sparse Decision Trees (OSDT) addresses
the myopic nature of conventional greedy tree induction by explicitly searching
for optimal sparse trees \citep{hu2019}. FIGS extends greedy tree learning by
simultaneously growing a flexible collection of trees, allowing compact
interpretable models to capture additive structure that can be inefficient for
a single tree to represent \citep{tan2025}.

A closely related perspective arises from subgroup discovery and contrast-set mining, which identify interpretable feature-defined regions exhibiting distinctive target distributions \citep{novak2009}. Such methods commonly construct increasingly specific rules by adding feature conditions, balancing subgroup coverage against selectivity. This provides a useful foundation for understanding hierarchical rule refinement, although the present work targets a different problem: constrained correction of residual prediction errors from a frozen classifier rather than discovery of descriptive subgroups.

The proposed method shares with this literature the use of simple,
interpretable feature-threshold rules, but differs in both objective and
deployment setting. It does not attempt to replace the original classifier
with an interpretable model. Instead, the original classifier remains frozen,
and the rule search operates only as a post-hoc correction mechanism. Moreover,
the graph search retains several candidate children at each level so that
alternative correction paths can be evaluated before a final path is selected.

\subsection{Post-hoc Classifier Correction and Patching}

The most closely related setting is classifier patching. Kauschke and
F{\"u}rnkranz \citep{kauschke2018} consider adaptation of a strong but
unchangeable black-box classifier by identifying regions of the input space in
which the original classifier is error-prone and learning a patch for those
regions. This shares an important motivation with our formulation: useful
corrections can be learned without retraining the underlying classifier.

Our setting differs in the structure and objective of the correction.
Rather than training an auxiliary classifier to model error regions, we search
for a compact conjunction of feature--threshold rules under an explicit
constraint on the adverse class-wise effect of the correction. Thus, the
search can directly target objectives such as removing FPs while limiting TP
loss, or recovering TPs while limiting additional FPs. The resulting
correction is a compact rule path that can be inspected independently of the
base classifier.

More broadly, recent post-hoc methods have investigated how pretrained
classifiers can be modified or supplemented without retraining their underlying
parameters. For example, post-hoc selective-classification methods modify the
confidence estimator associated with an existing classifier to improve its
risk--coverage behavior \citep{cattelan2024}. While these methods do not
perform the same form of label correction considered here, they reinforce the
practical motivation for improving the operational behavior of frozen
predictors through lightweight post-hoc mechanisms.

\subsection{Selective and Cost-Sensitive Classification}

Selective classification, also known as classification with a reject option,
allows a predictor to abstain on uncertain inputs and thereby trades coverage
against predictive risk \citep{elyaniv2010}. This idea has been extended to
deep neural networks, including post-hoc selective classifiers constructed
from already-trained networks \citep{geifman2017} and models such as
SelectiveNet that jointly optimize prediction and rejection
\citep{geifman2019}. The connection to our problem is the explicit control of
an undesirable outcome: both settings seek more useful operating points than
those obtained by optimizing aggregate classification accuracy alone.

There is nevertheless an important distinction. Selective classification
typically changes whether a prediction is issued, whereas our method changes
selected binary predictions through interpretable feature-dependent correction
rules. Consequently, our constraint is expressed directly in terms of changes
to FP and TP counts rather than through the conventional risk--coverage
trade-off.

Cost-sensitive classification provides another related perspective.
When false positives and false negatives have different consequences, the
optimal decision rule should account explicitly for those asymmetric costs
\citep{elkan2001}. Our formulation similarly recognizes asymmetric error
preferences, but expresses them as constrained post-hoc optimization. For
example, one operating mode searches for the largest reduction in FPs subject
to an upper bound on TP loss. This formulation is useful when the acceptable
degradation can be specified directly as an operational constraint rather than
as a single scalar misclassification cost.

\subsection{Motivation for Graph-based Correction Search}

These lines of research suggest three desirable properties for a post-hoc
correction mechanism: it should preserve the already-deployed base classifier,
produce a compact and inspectable correction, and explicitly control the
trade-off between corrected and newly introduced errors. Greedy tree induction
provides interpretability but can commit early to a locally attractive split;
optimal tree and rule-list methods address this limitation but generally seek
to learn an entire classifier. Selective and cost-sensitive methods explicitly
model prediction trade-offs but do not directly produce the feature-dependent
post-hoc correction path considered here.

Our graph-search formulation occupies the intersection of these settings. It
keeps the base classifier fixed, searches multiple alternative conjunctions of
feature--threshold conditions, and evaluates them according to a constrained
FP--TP objective. The dynamic optimization and reduced-histogram components
are introduced to make this multi-path search computationally practical.

```latex
\section{Method}

\subsection{Post-hoc Correction Formulation}

Let $C_{\theta}(x)$ denote a frozen binary classifier
with prediction $\hat{y}=C_{\theta}(x)$.
We seek a compact post-hoc correction rule that modifies
selected predictions without updating the base-model
parameters $\theta$.

A candidate correction path of depth $d\leq D$ is
parameterized by
\begin{equation}
\phi_d =
\big((j_1,\tau_1,s_1),\ldots,(j_d,\tau_d,s_d)\big),
\end{equation}
where $j_k$ is a feature index, $\tau_k$ is a threshold,
and $s_k\in\{\leq,>\}$ specifies the inequality direction.
The path defines a correction region
\begin{equation}
\mathcal{R}(\phi_d)=
\bigcap_{k=1}^{d}
\left\{x:x_{j_k}\;s_k\;\tau_k\right\}.
\label{eq:correction_region}
\end{equation}

Depending on the optimization mode, predictions satisfying
the rule are either changed from positive to negative
(FP reduction) or from negative to positive (TP recovery).
The correction is applied only to samples having the
corresponding original prediction.

The rule is learned from training features, ground-truth
labels, and base-model predictions $(X,y,\hat{y})$.
Once selected, the correction path is frozen and applied
to unseen samples without retraining $C_{\theta}$.

\subsection{Graph Search over Correction Rules}

The proposed graph search explores alternative correction
paths through successive feature-threshold conditions.
Each node represents a candidate path $\phi_d$, and
its children are generated by appending one additional
feature-threshold condition.

For a fixed path, successive correction regions are nested:
\begin{equation}
\mathcal{R}(\phi_{d+1})
\subseteq
\mathcal{R}(\phi_d).
\label{eq:nested_regions}
\end{equation}
Thus, each refinement restricts the samples eligible for
correction. This structure permits broad initial correction
regions to be progressively refined to reduce adverse
prediction changes.

At each node, candidate feature-threshold conditions are
evaluated using the correction objective defined below.
The top-$K$ candidates are retained as children, and
depth-first search (DFS) recursively explores the resulting
paths up to maximum depth $D$.

Unlike a conventional greedy decision tree, which commits
to a single locally preferred split at each node, the
proposed search retains multiple candidate refinements.
A rule that is not locally optimal may therefore lead to
a better complete correction path after further refinement.

The score of each node is evaluated over its complete
top-down rule path. After exploring the retained candidates,
the highest-scoring feasible path is selected:
\begin{equation}
\phi^* =
\arg\max_{\phi\in\mathcal{P}_{\mathrm{search}}}
J(\phi)
\quad
\text{subject to the correction constraint},
\end{equation}
where $\mathcal{P}_{\mathrm{search}}$ denotes the candidate
paths explored by the search procedure.

Only one correction path is selected during training and
applied to all eligible inference samples, rather than
using object-dependent paths as in a decision tree.

\subsection{Constrained Objective}

The objective is to maximize beneficial prediction
corrections while limiting adverse changes to predictions
that were originally correct.

For a candidate path $\phi_d$, define
$G(\phi_d)$ as the number of beneficial corrections
and $C(\phi_d)$ as the number of adverse corrections.

For TP recovery,
\begin{equation}
G(\phi_d)=\Delta TP(\phi_d),\qquad
C(\phi_d)=\Delta FP(\phi_d),
\end{equation}
where $\Delta TP$ counts recovered false negatives and
$\Delta FP$ counts newly introduced false positives.

For FP reduction,
\begin{equation}
G(\phi_d)=\Delta FP_{\mathrm{removed}}(\phi_d),
\qquad
C(\phi_d)=\Delta TP_{\mathrm{lost}}(\phi_d).
\end{equation}

Both modes can therefore be expressed using the
unified constrained objective
\begin{equation}
\max_{\phi} G(\phi)
\quad\text{subject to}\quad
C(\phi)\leq\epsilon,
\label{eq:constrained_objective}
\end{equation}
where $\epsilon$ is the maximum allowable adverse
correction cost for the selected mode.

To rank candidate paths during graph search, we use
the weighted objective
\begin{equation}
J(\phi)=\alpha'G(\phi)-C(\phi),
\label{eq:weighted_objective}
\end{equation}
where $\alpha'$ controls the relative importance of
beneficial and adverse corrections.
Candidate paths are ranked by $J(\phi)$, while feasibility
is determined by the corresponding correction constraint.

\paragraph{Objective improvement under rule refinement.}
For a fixed search path, let
$G_d=G(\phi_d)$ and $C_d=C(\phi_d)$.
The nested-region property in
Eq.~\ref{eq:nested_regions} implies
\begin{equation}
G_{d+1}\leq G_d,\qquad C_{d+1}\leq C_d.
\end{equation}

For a fixed weight $\alpha'$, the change in the
weighted objective after adding a feature-threshold
condition is
\begin{equation}
J(\phi_{d+1})-J(\phi_d)
=
(C_d-C_{d+1})
-\alpha'(G_d-G_{d+1}).
\label{eq:refinement_gain}
\end{equation}

Consequently, a refinement improves the objective if
\begin{equation}
C_d-C_{d+1}
>
\alpha'(G_d-G_{d+1}).
\label{eq:refinement_condition}
\end{equation}

This condition characterizes the trade-off underlying
progressive rule refinement: a new condition is beneficial
when the reduction in adverse corrections outweighs
the weighted loss of beneficial corrections.
Although deeper rules necessarily reduce correction
coverage, they may improve the overall objective by
eliminating adverse corrections more efficiently.

\subsection{Dynamic Optimization}

The nested structure of correction regions motivates
a depth-dependent constraint schedule. Rather than
enforcing the final adverse-correction budget at every
search depth, the proposed method permits broader
candidate regions at upper levels and progressively
tightens the constraint during refinement.

Let $\epsilon$ denote the final admissible adverse
correction budget. At depth $d$, the search uses
\begin{equation}
C(\phi_d)<\epsilon_d,\qquad
\epsilon_d=
\epsilon\left(\frac{D}{d+1}\right)^2,
\label{eq:dynamic_constraint}
\end{equation}
where $D$ is the maximum search depth and $d$ is
the current depth.

For TP recovery, $C(\phi_d)$ represents additional
false positives; for FP reduction, it represents
lost true positives. Thus, the same depth-dependent
schedule applies to both correction modes.

At shallow depths, the relaxed constraint allows
the search to retain regions with substantial
target-error coverage, even when their adverse
correction cost exceeds the final admissible budget.
As additional conditions restrict the correction
region, the allowable adverse correction cost
decreases.

The refinement condition in
Eq.~\ref{eq:refinement_condition} explains when
such specialization improves the weighted objective.
Together, multi-path exploration and dynamic
constraint tightening allow the algorithm to examine
promising intermediate candidates that might otherwise
be discarded under a fixed final constraint.

In the large example in Section~\ref{sec:large_case},
TP loss decreases from 14.20\% at depth 1 to
2.80\% at depth 3, while retaining a 17.52\%
FP reduction.
```

\subsection{Reduced Histogram Algorithm for Threshold Search}
At each node, features are ranked according to their optimal objective-function scores. The optimal score for each feature depends on its discrimination threshold. A direct threshold search can loop over all data points. The proposed histogram-and-bin algorithm instead searches over histogram bins. A finer bin size gives higher resolution but greater computational cost. To improve accuracy while reducing search cost, the method constructs a reduced histogram with variable bin sizes: coarse bins are used in regions dominated by either TP or FP, while fine bins are used in mixed regions where the discrimination threshold is likely to occur.

\begin{figure}[t]
\centering

\begin{minipage}[t]{0.48\linewidth}
\vspace{0pt}
\small
\textbf{Graph Search with Dynamic Optimization}

\vspace{3pt}
\hrule
\vspace{3pt}

\textbf{Input:} features $X$, labels $y$, base predictions $\hat{y}$,
maximum depth $D$, number of children $K$

\textbf{Output:} optimal correction path $P^*$

\vspace{3pt}

\textbf{1:} Initialize root node and empty path $P$

\textbf{2:} \textbf{for} depth $d=1,\ldots,D$ \textbf{do}

\textbf{3:}\quad Set dynamic TP/FP constraint for depth $d$

\textbf{4:}\quad \textbf{for each} candidate feature $x_j$ \textbf{do}

\textbf{5:}\qquad Build reduced TP/FP histogram

\textbf{6:}\qquad Search bins for best threshold

\textbf{7:}\qquad Determine threshold direction

\textbf{8:}\qquad Compute TP and FP variations

\textbf{9:}\qquad Compute correction score

\textbf{10:}\quad \textbf{end for}

\textbf{11:}\quad Rank candidate correction rules

\textbf{12:}\quad Keep top-$K$ rules as child nodes

\textbf{13:}\quad Expand child nodes recursively

\textbf{14:} \textbf{end for}

\textbf{15:} Rank complete top-down paths

\textbf{16:} Select highest-scoring feasible path $P^*$

\textbf{17:} \textbf{return} $P^*$

\vspace{3pt}
\hrule
\end{minipage}
\hfill
\begin{minipage}[t]{0.49\linewidth}
\vspace{0pt}
\begin{minipage}[t]{0.27\linewidth} 
\vspace{0pt}
\raggedright
\includegraphics[width=\linewidth]{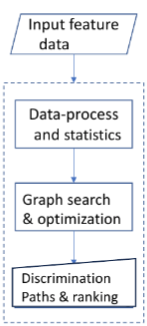}
\end{minipage}
\begin{minipage}[t]{0.53\linewidth} 
\vspace{0pt}
\raggedleft
\includegraphics[width=\linewidth]{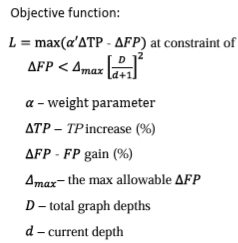}
\end{minipage}

\vspace{6pt}

\centering
\includegraphics[width=0.8\linewidth]{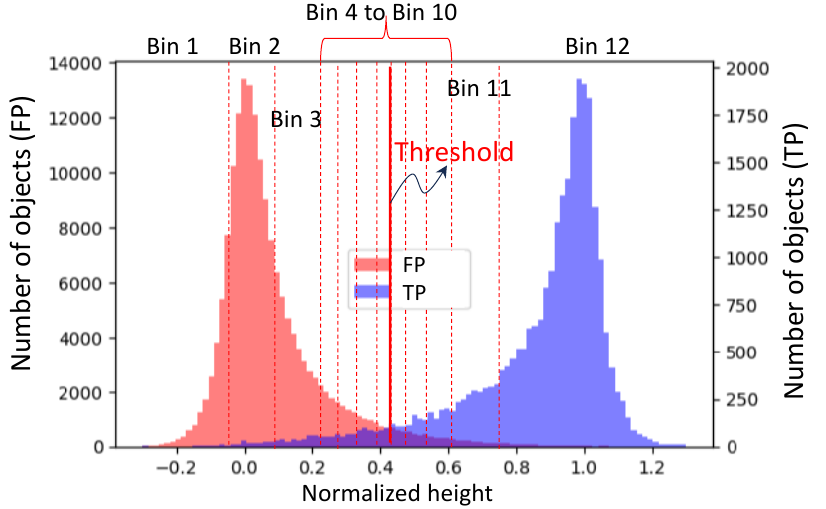}

\end{minipage}

\caption{
Overview of the graph-search algorithm with dynamic optimization and
reduced-histogram threshold search.
\textbf{Left:} Pseudocode of the complete graph-search procedure.
\textbf{Upper middle:} Flow chart of the search procedure.
\textbf{Upper right:} Dynamic optimization during graph search.
\textbf{Lower right:} Reduced-histogram representation for efficient
feature-threshold search.
}
\label{fig:algorithm_overview}

\end{figure}

The implementation of the proposed graph-search framework, including
dynamic optimization and reduced-histogram threshold search, is publicly
available.\footnote{\url{https://github.com/qinwuxutexas/class_discrimination_graph_search_dynamic_optimization}}

\section{Analysis and Results}
The following experiments demonstrate the behavior of the post-hoc correction method on predictions from a frozen SVM binary classifier. These examples illustrate the two constrained operating modes and the depth-dependent behavior of dynamic optimization; the broader multi-dataset evaluation protocol is specified in Appendix~\ref{app:openml_benchmark}.

\subsection{Illustrative Examples}

Examples 1 and 2 are small-scale illustrative cases intended to
visualize how the proposed graph-search correction modifies the
predictions of a frozen base classifier. They are not intended as
large-scale empirical evaluations. The large-scale evaluation is
presented separately in Example 3.

\paragraph{Example 1: False-positive reduction.}

\begin{figure}[t]
\centering
\includegraphics[width=.42\linewidth]{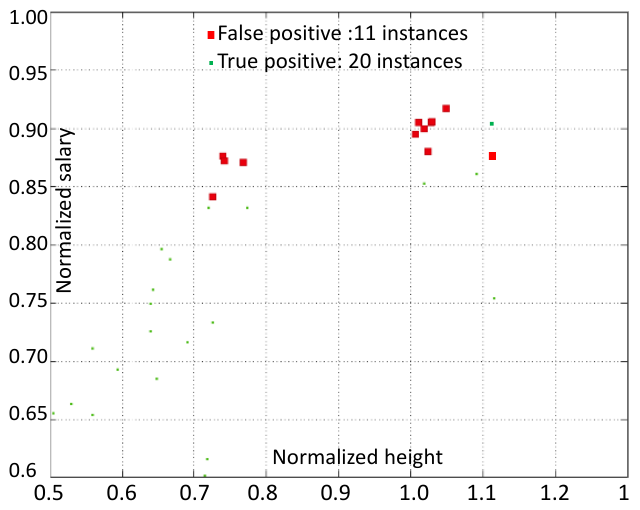}\hfill
\includegraphics[width=.48\linewidth]{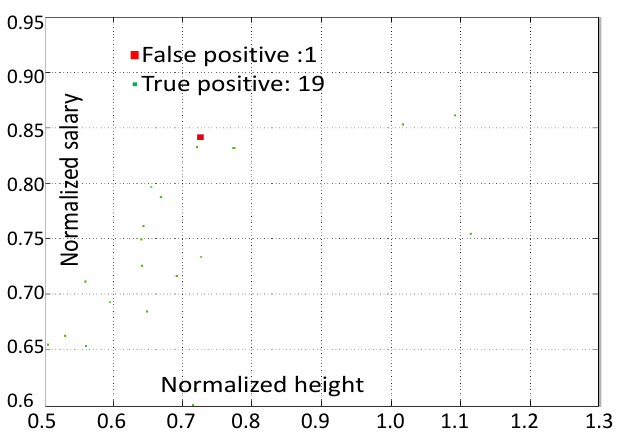}
\caption{FP reduction example. Left: base SVM alarms 11 FP and 20 TP objects. Right: the correction rule removes 10 FP while losing one TP.}
\label{fig:ex1}
\end{figure}

After applying an SVM model, 20 TP and 11 FP objects are alarmed. The goal is to reduce FP with TP loss $\leq5\%$. The graph search generates a rule with normalized salary threshold 0.865633. It clears 10 of 11 FP objects while losing one of 20 TP objects (5\%). Figure~\ref{fig:ex1} shows the base and corrected results.

\subsection{Example 2: TP Improvement, Small Case}
The SVM-predicted negative objects include 37 true negatives and 46 false negatives. The goal is to improve TP (rescue false negatives) with FP gain $\leq10\%$. The graph search classifies 30 additional objects as positive, including 27 false negatives rescued as TP and three true negatives converted to FP, corresponding to 59\% recovery of false negatives and an 8\% FP gain relative to the 37 true negatives.

\begin{figure}[t]
\centering
\includegraphics[width=.48\linewidth]{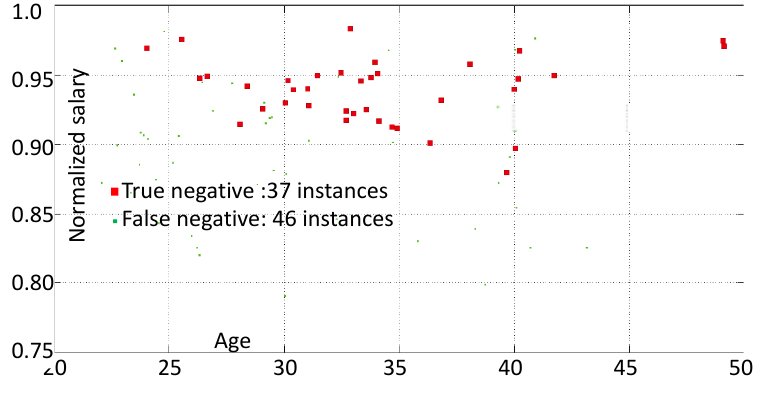}\hfill
\includegraphics[width=.48\linewidth]{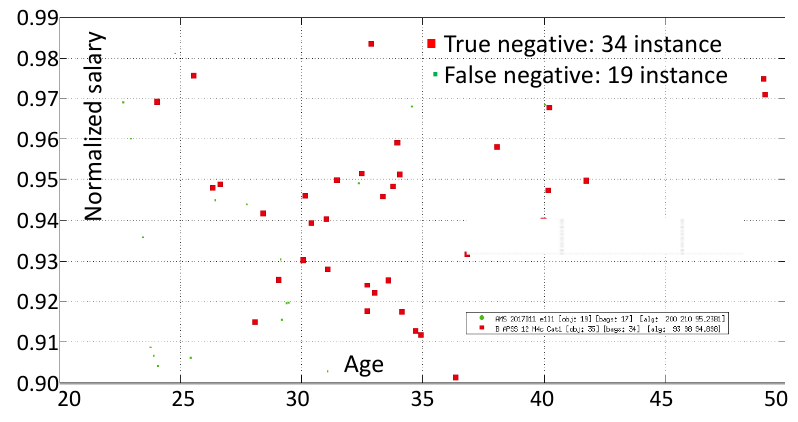}
\caption{TP improvement example. Left: base SVM negative predictions. Right: graph-search correction rescues 27 false negatives with three additional false positives.}
\label{fig:ex2}
\end{figure}

\subsection{Example 3: FP Reduction, Large Case}\label{sec:large_case}
The input contains 300,700 negative cases and 27,764 positive objects. The SVM predicts 1,884 FP. The goal is to reduce FP with TP loss $\leq3\%$. The graph search generates a three-level discrimination path using dynamic optimization. At depth 1 it reduces FP by 43.52\% but incurs 14.20\% TP loss. At deeper levels both FP reduction and TP loss decrease, resulting in a final FP reduction of 17.52\% with TP loss of 2.80\%, satisfying the constraint.

\begin{table}[t]
\centering
\caption{FP reduction and TP loss at three graph-search levels.}
\label{tab:large}
\begin{tabular}{rrrrrr}
\toprule
Node & Depth & $\Delta$FP & FP reduction & $\Delta$TP & TP loss\\
\midrule
2 & 1 & 820 & 43.52\% & 259 & 14.20\%\\
7 & 2 & 607 & 32.22\% & 178 & 9.76\%\\
28 & 3 & 330 & 17.52\% & 51 & 2.80\%\\
\bottomrule
\end{tabular}
\end{table}



\section{Discussion}
The method is intended for settings in which a deployed or otherwise fixed classifier exhibits structured residual errors that can be characterized by interpretable input features. Its principal advantage is that it does not require modifying or retraining the base model: the correction path is learned separately and can be inspected before deployment. The constrained formulation also exposes the operational trade-off between errors corrected and new errors introduced.

The method has corresponding limitations. Because the correction path is based on feature thresholds, it is most appropriate when meaningful structured features are available and residual errors contain repeatable structure. A compact rule cannot be expected to correct errors that are effectively random with respect to the available features. Search depth, branch factor, histogram resolution, and the constraint schedule also affect the search space and computational cost. Appendix~\ref{app:openml_benchmark} is designed to test whether gains persist across datasets and frozen base classifiers rather than only on the original SVM example.

\section{Conclusion}
We presented a model-agnostic framework for constrained post-hoc error correction in binary classifiers. The method leaves the base classifier frozen and searches for a compact feature--threshold rule path that either removes false positives subject to a bound on true-positive loss or recovers true positives subject to a bound on false-positive gain. Graph search retains multiple candidate paths, dynamic optimization progressively enforces the target constraint with depth, and reduced-histogram threshold search improves computational efficiency. The current examples demonstrate both correction modes, including 90\% FP removal at 5\% TP loss in the small case and 17.52\% FP reduction at 2.80\% TP loss in the large case. A public multi-dataset benchmark is specified in Appendix~\ref{app:openml_benchmark} to evaluate generalization, constrained error trade-offs, and computational efficiency across stronger frozen baselines.

\bibliographystyle{plainnat}
\bibliography{main}

\appendix
\section{Software Architecture and Reproducibility}

We provide a public Python implementation of the proposed graph-search
framework.\footnote{\url{https://github.com/qinwuxutexas/class_discrimination_graph_search_dynamic_optimization}}
The released code separates data processing, correction objectives,
rule representation, reduced-histogram threshold search, and graph search
into modular components.

Given feature vectors $X$, ground-truth labels $y$, and predictions
$\hat{y}$ from a frozen base classifier, the implementation first loads
and preprocesses the input data. Candidate feature--threshold rules are
then evaluated using reduced-histogram threshold search. At each graph
depth, candidate rules are ranked according to the correction
objective, and the top candidates are retained as child nodes. Dynamic
optimization adjusts the admissible TP/FP constraint as the search
progresses. The resulting top-down paths are ranked, and the
highest-scoring feasible path is selected.


In the reported large case, the implementation generated 39 discrimination rules in 9 seconds for 300,700 negative and 27,764 positive input objects on a dual-core 2.59 GHz laptop, with maximum process memory of 122 MB.

The learned path is fitted using the training data and is subsequently
frozen and applied unchanged to the held-out test predictions. The
repository includes configuration files, executable scripts, the core
search implementation, data utilities, and unit tests. Instructions for
installation and reproduction are provided in the repository README.

\section{Proposed OpenML Benchmark Evaluation}
\label{app:openml_benchmark}

To further evaluate the generality of the proposed graph-search method,
we plan a large-scale benchmark using public binary-classification tasks
from OpenML. The purpose of this experiment is to evaluate whether the
proposed method can improve already-trained classifiers on datasets for
which substantial classification errors remain, rather than selecting
datasets based on the observed improvement of our method.

\subsection{Dataset Selection}

We use the OpenML-CC18 benchmark suite as the initial candidate pool.
The suite contains 72 curated classification tasks covering datasets
with different sample sizes, feature dimensions, class distributions,
and levels of classification difficulty.

The benchmark datasets will be selected \emph{before} applying the
proposed graph-search algorithm according to the following procedure:

\begin{enumerate}
    \item Retrieve binary-classification tasks from the OpenML-CC18
    benchmark suite.

    \item Retrieve existing OpenML runs for standard classifiers,
    primarily SVM and tree-based classifiers such as XGBoost or
    gradient-boosted trees.

    \item Extract the reported test performance, model configuration,
    preprocessing pipeline, and hyperparameters for reproducible runs.

    \item Rank candidate datasets according to the performance of the
    existing base classifiers.

    \item Retain approximately eight datasets for which strong
    conventional classifiers still exhibit substantial residual
    classification error.

    \item Fix this dataset set before evaluating the proposed
    graph-search correction method.
\end{enumerate}

This screening procedure deliberately uses only the performance of the
\emph{base classifier}. No result from the proposed graph-search method
is used for dataset selection. This prevents selecting datasets
specifically because they favor the proposed method.

\subsection{Reproduction of Base Classifiers}

For each selected dataset, we first identify a strong and reproducible
OpenML run. OpenML runs provide the model flow, hyperparameters,
evaluation results, and, when available, prediction outputs. These
artifacts are used to reproduce the base classifier as closely as
possible.

The following hierarchy is used:

\begin{enumerate}
    \item If a suitable OpenML run provides the required predictions,
    dataset split information, and model configuration, its predictions
    are used directly as the frozen base-classifier output.

    \item If predictions are unavailable but the model architecture,
    preprocessing procedure, and hyperparameters are available, the
    corresponding classifier is retrained using the reported
    configuration and the OpenML task splits.

    \item If an existing run cannot be reliably reproduced, a strong
    SVM or XGBoost baseline is trained and tuned using a predefined
    hyperparameter search procedure.
\end{enumerate}

The fitted weights of the original OpenML model are therefore not
required. What is required is either the original predictions or
sufficient information to reproduce a comparable base classifier.

\subsection{Post-hoc Graph-Search Evaluation}

Let the reproduced base classifier be denoted by
$C_{\theta}$, where $\theta$ remains fixed during the proposed
post-hoc optimization.

For each benchmark dataset, the experiment proceeds as follows:

\begin{enumerate}
    \item Train or reproduce the base classifier $C_{\theta}$.

    \item Obtain its predictions on the training data and identify the
    residual false-positive and false-negative samples.

    \item Apply the proposed graph-search algorithm to
    $(X_{\mathrm{train}},y_{\mathrm{train}},
    \hat{y}_{\mathrm{train}})$.

    \item Search feature--threshold correction rules using the
    reduced-histogram procedure.

    \item Construct candidate top-down paths using graph search with
    dynamic optimization.

    \item Select the highest-scoring feasible path according to the
    specified TP/FP constraint.

    \item Freeze the resulting discrimination path.

    \item Apply exactly the same path, without further optimization,
    to the held-out test predictions of $C_{\theta}$.

    \item Compare the original classifier $C_{\theta}$ against
    $C_{\theta}+\mathrm{GraphSearch}$ on exactly the same test samples.
\end{enumerate}

Thus, the proposed method does not retrain or modify the parameters of
the underlying classifier. It learns a constrained post-hoc correction
rule from residual classification errors and subsequently applies the
frozen rule to unseen samples.

\subsection{Candidate Dataset Screening}

The initial screening will focus on binary datasets satisfying the
following criteria:

\begin{itemize}
    \item sufficient sample size to support separate training and
    held-out evaluation;
    \item nontrivial residual error for strong SVM or boosted-tree
    classifiers;
    \item numerical or consistently encoded tabular features suitable
    for threshold-based discrimination rules;
    \item reproducible OpenML tasks and model runs; and
    \item sufficient false-positive and/or false-negative samples to
    evaluate constrained error correction.
\end{itemize}

Candidate datasets will first be ranked using existing OpenML
performance records. Approximately eight difficult datasets will then
be frozen as the benchmark set before graph-search experiments are
performed.

\subsection{Evaluation Protocol}

For every dataset, the principal comparison is

\[
C_{\theta}
\quad\mathrm{vs.}\quad
C_{\theta}+\mathrm{GraphSearch},
\]

where both methods use the same base model, data split, preprocessing,
and test samples. We report accuracy, precision, recall, F1 score,
false-positive rate, false-negative rate, and the absolute numbers of
corrected and newly introduced errors.

Because the method is designed for constrained error correction, we
additionally report the trade-off between recovered errors and the
errors introduced by the correction path. For example, in the
false-positive-reduction setting, the primary quantities are the
percentage of false positives removed and the associated loss of true
positives.

Runtime, number of evaluated candidate rules, selected path depth, and
the final rule path are also reported for each dataset. These
measurements evaluate both the effectiveness and computational
efficiency of the proposed search procedure.

\end{document}